%% file: main.tex
\documentclass[10pt,twocolumn,letterpaper]{article}

\usepackage{cvpr}              


\usepackage{xspace}
\usepackage{bbding}
\usepackage{multirow}
\usepackage{pifont}

\newcommand{\First}{{\em First}}
\newcommand{\myie}{{\em i.e.}}
\newcommand{\Second}{{\em Second}}
\newcommand{\system}{{StreamingAssistant}\xspace}
\newcommand{\para}[1]{\vspace{1pt}  \noindent {\bf #1}}

\usepackage[table]{xcolor} 

\definecolor{cvprblue}{rgb}{0.21,0.49,0.74}
\usepackage[pagebackref,breaklinks,colorlinks,allcolors=cvprblue]{hyperref}

\def\paperID{} 
\def\confName{CVPR}
\def\confYear{2026}

\title{\system: Efficient Visual Token Pruning for \\ Accelerating Online Video Understanding}

\author{
Xinqi Jin$^{1,3,*}$\ \ \ \ \ \ 
Hanxun Yu$^{2,*}$\ \ \ \ \ \ 
Bohan Yu$^3$\ \ \ \ \ \ 
Kebin Liu$^{1,\text{\Envelope}}$\ \ \ \ \ \ 
Jian liu$^{3,\text{\Envelope}}$\ \ \ \ \ \ 
Keda Tao$^{2,3,4}$
\\
Yixuan Pei$^3$\ \ \ \ \ \ 
Huan Wang$^4$\ \ \ \ \ \ 
Fan Dang$^5$\ \ \ \ \ \ 
Jiangchuan Liu$^6$\ \ \ \ \ \ 
Weiqiang Wang$^3$\ \ \ \ \ \ 
\vspace{8pt}\\  
$^1$Tsinghua University\ \ \ \ \ \  $^2$Zhejiang University\ \ \ \ \ \  $^3$Ant Group 
\\$^4$Westlake University\ \ \ \ \ \  $^5$Beijing Jiaotong University\ \ \ \ \ \  $^6$Simon Fraser University
}

\begin{document}

\definecolor{gray0}{gray}{0.9}
\definecolor{gray1}{gray}{0.8}
\definecolor{gray2}{gray}{0.7}

\maketitle

\renewcommand{\thefootnote}{*}
\footnotetext{Co-first authors}

\renewcommand{\thefootnote}{\Envelope}
\footnotetext{Co-corresponding authors (kebinliu2021@tsinghua.edu.cn, rex.lj\\@antgroup.com)}

\input{sec/0_abstract}    
\input{sec/1_intro}
\input{sec/2_related_work}
\input{sec/3_methods}
\input{sec/4_experiment}
\input{sec/5_future_work}
\input{sec/6_conclusion}
{
    \small
    \bibliographystyle{ieeenat_fullname}
    \bibliography{main}
}


\end{document}

%% file: sec/0_abstract.tex
\begin{abstract}
Online video understanding is essential for applications like public surveillance and AI glasses. However, applying Multimodal Large Language Models (MLLMs) to this domain is challenging due to the large number of video frames, resulting in high GPU memory usage and computational latency. To address these challenges, we propose token pruning as a means to reduce context length while retaining critical information. Specifically, we introduce a novel redundancy metric, Maximum Similarity to Spatially Adjacent Video Tokens (MSSAVT), which accounts for both token similarity and spatial position. To mitigate the bidirectional dependency between pruning and redundancy, we further design a masked pruning strategy that ensures only mutually unadjacent tokens are pruned. 
We also integrate an existing temporal redundancy-based pruning method to eliminate temporal redundancy of the video modality.
Experimental results on multiple online and offline video understanding benchmarks demonstrate that our method significantly improves the accuracy (\myie, by 4\% at most) while incurring a negligible pruning latency (\myie, less than 1ms).
Our full implementation will be made publicly available.

\end{abstract}

%% file: sec/1_intro.tex
\section{Introduction}
\label{sec:intro}

Online video understanding is a critical technique in many real-world applications, such as public surveillance~\cite{fontes2022ai,wei2023moire} and AI glasses~\cite{waisberg2024meta,rossos2024ai}. 
With the recent advancements in Multimodal Large Language Models (MLLMs), significant efforts~\cite{llava-ov,wang2024qwen2-vl,bai2025qwen2.5-vl,team2024gemini1.5} have been devoted to extending their impressive interaction capabilities to vision-language tasks.
However, in online video understanding applications, video frames are continuously generated, and even a single frame is encoded into a large number of video tokens (\eg hundreds or even thousands of tokens per frame).
Such an excessively long input sequence incurs a heavy GPU memory overhead as well as a huge response latency that harms the user's interaction experience.

To effectively reduce the input sequence, one practical method is token pruning.
The rationale behind token pruning lies in the high redundancy of the video modality.
One major kind of redundancy is spatial redundancy.
Existing works typically pinpoint spatially redundant tokens through two methods, both of which have several drawbacks:

\begin{itemize}
    \item{\textbf{Attention-based methods.}}
    Attention-based methods~\cite{yang2025visionzip, shang2024LLaVA-PruMerge} quantify the importance of a video token via its received attention scores in the video encoder.
    However, tokens receiving high attention weights are typically important to the training task and data, but not necessarily important to the user's query.
    Furthermore, efficient attention implementations~\cite{dao2022flashattention} may not compute the attention scores explicitly, so these methods require slow and memory-intensive attention implementations.
    \item{\textbf{Similarity-based methods.}}
    Other works~\cite{divprune, bolya2022tome, wang2025folder} quantify the redundancy of a token through its similarity to other video tokens.
    Intuitively, a token can be highly redundant if it is very similar to another video token.
    However, these tokens are represented by not only their embeddings but also their spatial indices, with the latter being modeled by the positional embedding module in the Large Language Model (LLM) part of the MLLM.
    Even if a token is similar to another remotely located token, it may not be redundant due to its significantly different positional embedding.
    Simply ignoring this issue can lead to the blurring or loss of spatial position information.
\end{itemize}

\begin{figure*}[tp]
    \centering
    \includegraphics[width=0.8\linewidth]{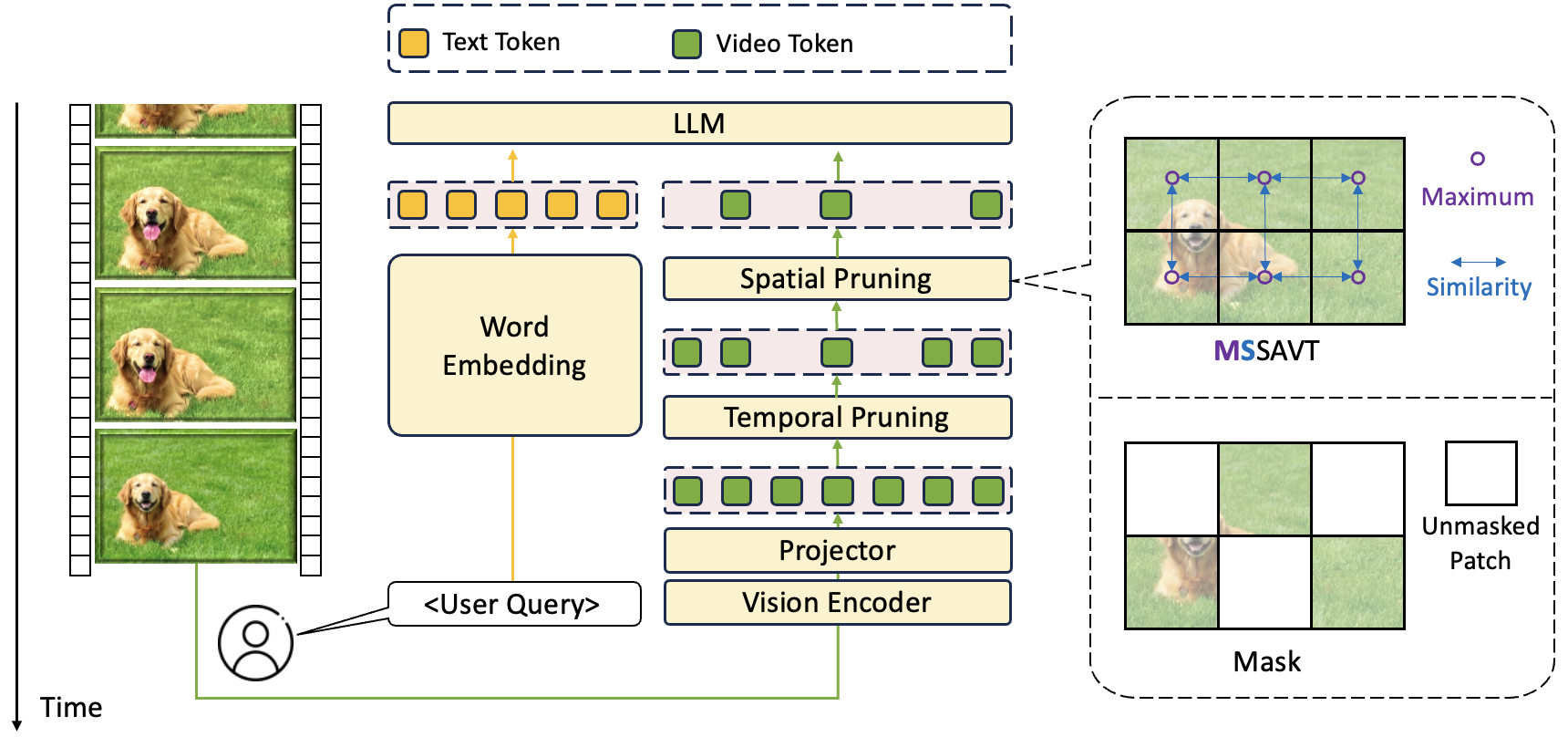}
   \caption{We introduce \system, a token pruning method to eliminate both temporal and spatial redundancy in the video tokens.
   Its key innovation lies in the novel spatial redundancy metric (MSSAVT) and the masked pruning strategy to disentangle pruning from redundancy quantification.}
    \label{fig:intro}
\end{figure*}

In \system, we propose a novel redundancy metric - \textbf{M}aximum \textbf{S}imilarity to \textbf{S}patially \textbf{A}djacent \textbf{V}ideo \textbf{T}okens (MSSAVT) to address the aforementioned drawbacks.
While the MSSAVT metric remains within the spectrum of similarity-based approaches, its key advancement lies in restricting the comparison to spatially adjacent tokens.
This design, rather than conducting a global similarity comparison to other tokens, ensures that the redundancy quantification inherently preserves and is guided by the tokens' positional context.

Despite offering a more principled characterization of spatially redundant regions, strictly using the MSSAVT metric as the pruning standard is still challenging.
We observe a critical and complex entanglement effect between token pruning and redundancy quantification.
Specifically, while token pruning operates under the guidance of precomputed redundancy values, this relationship is inherently bidirectional.
The redundancy score of a token can be dynamically altered by the pruning of its spatial neighbors — removing one token may also reduce the redundancy values of its spatially adjacent tokens.
The interleaving execution of pruning and redundancy update solves this issue at the cost of a large pre-processing latency, while simply ignoring this issue can lead to over-pruning and the erroneous removal of essential tokens.
To break this entanglement effect while still enabling parallel and fast pruning, we propose a masked pruning strategy.
The mask is carefully designed, and we only regard tokens selected by the mask as pruning candidates.
We prove that this pruning strategy incurs neither over-pruning nor missed pruning.

We also consider the temporal redundancy in \system.
As shown in Fig.~\ref{fig:intro}, we first use temporal pruning before applying spatial pruning.
We currently use the DTD algorithm proposed in TimeChat-Online~\cite{timechatonline} for temporal pruning, but other temporal pruning methods can be used as well.
A token is finally retained only if it is both temporally and spatially non-redundant.

We evaluate our method on multiple widely used online and offline video understanding benchmarks.
Under the same pruning ratio, \system achieves a higher response accuracy than many baseline methods.
For example, it improves the accuracy by 4\% at most compared to the DTD baseline.
The pruning process of \system lasts less than 1ms and has a negligible impact on the Time to First Token (TTFT) latency.
We also conduct comprehensive ablation studies to demonstrate the benefits of the MSSAVT metric and the masked pruning strategy.

To summarize, our contributions are as follows:
\begin{itemize}
    \item To preserve the crucial positional information, we propose the novel MSSAVT metric to quantify spatial redundancy in video tokens.
    \item We identify and address the entanglement effect between token pruning and redundancy quantification.
    To break this entanglement while enabling efficient parallel processing, we introduce a masked pruning strategy whose reliability is theoretically guaranteed.
    \item We conduct extensive experiments on multiple video understanding benchmarks. 
    \system achieves superior accuracy under the same pruning ratio while incurring a negligible pre-processing latency.
\end{itemize}

\begin{figure*}[tp]
    \centering
    \includegraphics[width=\linewidth]{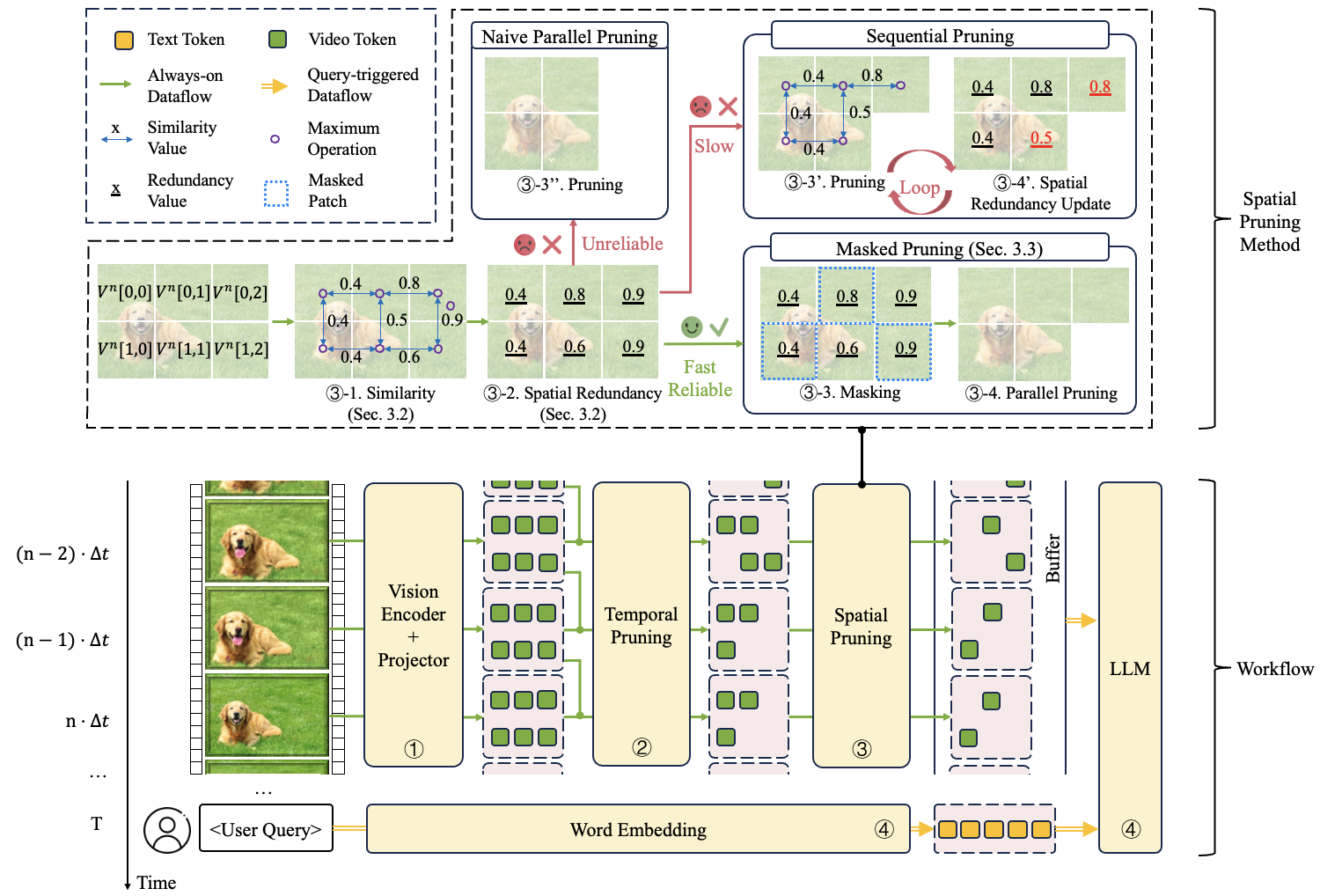}
   \caption{\textbf{Bottom}: the workflow of \system. \textbf{Top}: the proposed spatial pruning method along with several sub-optimal methods.}
    \label{fig:workflow}
\end{figure*}

%% file: sec/2_related_work.tex
\section{Related Work}
\label{related_work}

\textbf{Online/Streaming Video Understanding.} 
The advancement of MLLMs~\cite{lin2023video-llava,team2024gemini1.5,zhang2025videollama,yu2025inst3d,bai2025qwen2.5-vl,llava-ov} has spurred numerous efforts to leverage their powerful understanding and generation capabilities for video-related tasks. Among these, VideoLLM-Online~\cite{chen2024videollm-online} pioneers the development of MLLMs for streaming video understanding, which introduces a streaming EOS prediction mechanism, enabling the model to learn when to respond or remain silent. However, its performance is constrained by the limited per-frame visual token input, stemming from the lack of effective streaming context compression. Subsequent research has focused on efficiently encoding dense video streams. For example, VideoLLM-MoD~\cite{wu2024videollm-mod} addresses this limitation by incorporating a mixture of depth to optimize visual token computation, thus enabling higher input resolution. ReKV~\cite{di2025rekv} and Inf-MLLM~\cite{ning2024inf} focus on improving KV Cache management for better video context handling. StreamChat~\cite{liu2024streamchat} and VideoChat-Online~\cite{huang2025videochat-online} employ dynamic memory banks to retain informative video tokens. 
In contrast, our approach takes a more comprehensive view by addressing both temporal and spatial redundancy, proposing a novel masked pruning strategy that not only improves accuracy but also boosts overall performance.

\noindent\textbf{Visual Token Pruning for MLLMs.} The computational complexity of Transformer-based networks scales with the input token length, and MLLMs typically contain more visual tokens than textual ones. Existing methods for reducing spatial redundancy include VisionZip~\cite{yang2025visionzip} and LLaVA-PruMerge~\cite{shang2024LLaVA-PruMerge}, which quantify token importance based on attention scores from the CLS token or other video tokens. FastV~\cite{chen2024fastv} prunes redundant tokens at specific layers of LLM based on the task-oriented attention importance. ToMe~\cite{bolya2022tome} and FOLDER~\cite{wang2025folder} quantify the redundancy of a token through its similarity to other video tokens. Other approaches target reducing the temporal redundancy in videos. TempMe~\cite{shen2024tempme} minimizes temporal redundancy by merging adjacent video clips, while DTD~\cite{timechatonline} prunes each incoming frame independently and adaptively drops video tokens to reduce the computational burden. HoliTom~\cite{holitom} tackles temporal redundancy through global redundancy-aware video segmentation. In this paper, we adopt the DTD pruning algorithm from TimeChat-Online~\cite{timechatonline} for its simplicity and effectiveness. Another line of work~\cite{zhang2024sparsevlm,shen2024longvu} also leverages language guidance through user queries for token pruning. However, they require reprocessing all previous frames for each new query and are prone to noisy or misleading textual clues. In contrast, we avoid language guidance and utilize the MSSAVT metric to efficiently quantify each token's spatial redundancy and prune the most redundant tokens.

%% file: sec/3_methods.tex
\section{StreamingAssistant}
\label{sec:system}

\subsection{System Workflow}
\label{sec:system-workflow}

The workflow of \system is shown in Fig.~\ref{fig:workflow}.
Video frames are continuously fed to \system, with the interval between two consecutive frames being $\Delta t$.
In contrast, the user query may be sent at any timestamp.
Consequently, there are two categories of dataflows in \system, including the always-on dataflow and the user query-triggered dataflow.
The always-on dataflow is triggered at $n\Delta t$ for every integer $n$.
In the always-on dataflow, the received video frame is first transformed into video tokens by the vision encoder and projector parts of the MLLM (corresponding to Step \ding{192} in Fig.~\ref{fig:workflow}).
To reduce the number of video tokens and improve the MLLM's generation efficiency, the video tokens are successively processed by the temporal pruning module (Step \ding{193}) and the spatial pruning module (Step \ding{194}).
The retained tokens are stored in the buffer.
The user query-triggered dataflow is triggered at time $T$, when the user asks a question about the past video stream.
The query is converted into text tokens through the processing of the word embedding layer.
The text tokens and video tokens currently in the buffer are together fed to the Large Language Model (LLM) part of the MLLM to produce the response (Step \ding{195}).

At the core of \system is its token pruning strategy.
Specifically, the frame received at $n\Delta t$ is converted into a matrix of video tokens $V^n\in \mathbb{R}^{W\times H\times D}$, with $W$ (or $H$) denoting the number of tokens per row (or column) and $D$ denoting the embedding dimension.
The temporal pruning module in \system uses the DTD algorithm, which is detailed in the paper of TimeChat-Online~\cite{timechatonline}.
The temporal pruning module compares $V^n$ with $V^{n-1}$ to produce a boolean dropping mask $M^n_t\in \{True, False\}^{W\times H}$, with $True$ indicating dropping and $False$ indicating no dropping.
The spatial pruning module takes $V^n$ as input and also produces a boolean dropping mask $M^n_s\in\{True, False\}^{W\times H}$.
The details of the proposed spatial pruning algorithm are discussed later in Sec.~\ref{sec:system-quantification} and Sec.~\ref{sec:system-mask}.
A token will be retained only if it's neither temporally redundant nor spatially redundant.
Or, formally speaking, the final dropping mask $M^n$ is computed as $M^n_{t} \ | \ M^n_{s}$.

\subsection{Quantification of Spatial Redundancy}
\label{sec:system-quantification}
For spatial redundancy-based pruning, we decide whether to prune a token based on whether its spatial redundancy metric exceeds a threshold $\tau_s$.
Therefore, the design of the spatial redundancy metric has a great impact on the performance of \system.

We notice that existing works typically quantify spatial redundancy via either the attention weights within the vision encoder~\cite{yang2025visionzip, holitom, shang2024LLaVA-PruMerge} or the similarity between tokens~\cite{divprune, bolya2022tome, wang2025folder}.
We do not consider the attention weights-based methods for two reasons.
\First, these methods may be affected by the discrepancy between the training objective and the actual usage scenario.
For example, when the vision encoder has a CLS token (such as the CLIP encoder~\cite{clip}), these methods quantify a video token's redundancy as the CLS token's attention weight to this video token.
However, the CLS token is typically used for classification in the training stage, so the CLS token's attention weight to some video token only indicates how much information related to the classification task is included in the video token.
Yet, the user query's task may differ greatly from the classification task, making the CLS token-guided quantification method unreliable.
\Second, efficient attention implementations (such as FlashAttention~\cite{dao2022flashattention}) do not directly compute the attention scores, while naive implementations can cause problems such as high computation latency and significant memory overhead.
Therefore, attention weights-based methods are inherently more difficult to deploy.

Different from attention weights-based methods, similarity-based methods directly derive the redundancy of a token by its maximum similarity to other tokens in the frame.
The intuition is that a token provides almost no new information if it is highly similar to other tokens.
This category of methods is superior in terms of less dependency on the training objective and better compatibility with efficient attention implementations.
Despite the advantages, these methods are still sub-optimal due to their neglect of the position information.
While feeding a video token to the LLM part of the MLLM, both its feature embedding and positional index are required by the LLM.
The LLM uses positional embedding techniques (such as the Multimodal Rotary Position Embedding~\cite{RoPE, wang2024qwen2-vl, bai2025qwen2.5-vl} to encode the positional information into the token's feature.
From the perspective of the LLM, a token can still be informative even if its feature embedding is similar to a spatially distant token.
As existing methods neglect this problem, they can cause over-pruning of informative tokens and losses or blurrings of the positional information.

To further improve the similarity-based methods, we propose to derive the redundancy values from the \textbf{M}aximum \textbf{S}imilarity to \textbf{S}patially \textbf{A}djacent \textbf{V}ideo \textbf{T}okens (MSSAVT).
Formally, we denote the token corresponding to the patch located at the $i$-th row and the $j$-th column of the $n$-th frame as $V^n[i,j]\in\mathbb{R}^d$.
Its spatial redundancy is computed as 
\begin{equation}
  \label{eq:spatial_redundancy}
\begin{aligned}
  R^n_s[i,j] = & max_{(\delta_i, \delta_j)\in\{(-1,0),(1,0),(0,-1),(0,1)\}}\Big\{\\
  &Sim(V^n[i,j], V^n[i+\delta_i, j+\delta_j])\Big\},
\end{aligned}
\end{equation}
with $Sim(\cdot, \cdot)$ denoting the cosine similarity between two token embeddings.
We limit the value of $\delta_i$ and $\delta_j$ so that only the similarities to spatially neighbouring tokens are taken into consideration.
This subtle design fundamentally changes the redundancy definition from a global, position-agnostic comparison to a local, position-aware one.
By definition, the positional context of a token is captured by the identities of its neighbors. Therefore, MSSAVT intrinsically incorporates positional information into the redundancy score, directly addressing the core limitation of previous similarity-based methods.

The computation process is also visualized in the top part of Fig.~\ref{fig:workflow}.
The similarity between adjacent tokens is computed in step \ding{194}-1 and the MSSAVT metric is obtained in step \ding{194}-2.
We also quantitatively demonstrate the benefits of the MSSAVT metric in Sec.~\ref{sec:exp-ablation}.

\subsection{Masked Pruning Strategy}
\label{sec:system-mask}

Intuitively, token pruning methods rely on the pre-computed redundancy values to perform selective pruning.
However, the token pruning process can also affect the redundancy values.
An example is shown in Fig.~\ref{fig:workflow}.
Initially, in step \ding{194}-2 of the figure, the token $V^n[1,2]$ has a high MSSAVT value of $R^n_s[1,2]=0.9$.
When $V^n[1,2]$ is pruned in step \ding{194}-3', it does not contribute to the spatial redundancy of its adjacent tokens (\ie $V^n[0,2]$ and $V^n[1,1]$) anymore.
Consequently, in step \ding{194}-4', $R^n_s[0,2]$ decreases from $0.9$ to $0.8$, and $R^n_s[1,1]$ also decreases from $0.6$ to $0.5$.
This example demonstrates the complex entanglement between token pruning and spatial redundancy.
If we strictly use the MSSAVT metric as our pruning standard, we have to iteratively perform token pruning (\ie, step \ding{194}-3') and redundancy update (\ie, step \ding{194}-4').
Such an interleaving and sequential pruning strategy can lead to a large pre-processing latency, which constitutes a part of the TTFT latency and offsets token pruning's benefits to some extent.

Surely, we can simply ignore the entanglement issue and directly drop all the tokens with a large initial MAASVT value.
Although this naive parallel pruning strategy incurs a minimal pre-processing latency, it may cause over-pruning and the loss of necessary information.
For example, let's assume a spatial dropping threshold of $0.85$.
In such a setting, the naive parallel strategy simultaneously drops $V^n[0,2]$ and $V^n[1,2]$ (as shown in step \ding{194}-3'' of Fig.~\ref{fig:workflow}).
However, as shown in step \ding{194}-4', $V^n[0,2]$ becomes less redundant and has a redundancy value below the threshold $0.85$ after the dropping of $V^n[1,2]$.

To handle the entanglement between pruning and redundancy without compromising the pre-processing latency, we propose a masked pruning strategy.
The mask can be viewed as a boolean matrix $M_p\in\{True, False\}^{W\times H}$, with its element $M_p[i,j]$ set as 

\begin{equation}
\label{eq:mask}
M_p[i,j]=\left\{
\begin{aligned}
True & , & if\ (i+j)\%2 == 1, \\
False & , & otherwise,
\end{aligned}
\right.
\end{equation}
As shown in step \ding{194}-3 of Fig.~\ref{fig:workflow}, only tokens selected by the mask $M_p$ (\ie tokens whose corresponding element in the mask is $True$) are considered as candidates for pruning.
In step \ding{194}-4, the strategy drops redundant tokens with the spatial dropping mask
\begin{equation}
\label{eq:mns}
M^n_s=(R_s>\tau_s)\ \&\ M_p.
\end{equation}

Intuitively, our masked pruning strategy enables parallel pruning and incurs negligible pre-processing latency.
We further discuss the benefits of the masked pruning strategy by answering the following two questions:

\para{Question \#1: does masked pruning avoid over pruning?}

\para{Answer \#1:}
\system always obeys the rule of dropping tokens of large MSSAVT values, even if it never updates the MSSAVT value.
Formally, if a token $V^n[i,j]$ is a pruning candidate selected by the mask $M_p$, we have $(i+j)\%2 == 1$.
According to the definition of MSSAVT, the pruning of $V^n[i,j]$ will only affect the MSSAVT metrics of its four neighbouring tokens $V^n[i-1,j]$, $V^n[i+1,j]$, $V^n[i,j-1]$, $V^n[i,j+1]$.
As 
\begin{equation}
\label{eq:prove0}
\begin{aligned}
    &((i-1)+j)\%2 \\
  = & ((i+1)+j)\%2 \\
  = & (i+(j-1))\%2 \\
  = & (i+(j+1))\%2 \\
  = & 0,
\end{aligned}
\end{equation}
none of the four tokens is also the pruning candidate selected by $M_p$.
Therefore, the pruning of $V^n[i,j]$ will not affect the MSSAVT metric of any other pruning candidates $V^n[i',j']$.
\system does not need to update $R^n_s[i',j']$ as it will not decrease at all.
No over-pruning will occur due to the omission of unnecessary MSSAVT updates.

\para{Question \#2: is it possible that masked pruning misses potential pruning chances?}

\para{Answer \#2:}
We point out that tokens not selected by the mask $M_p$ must be unredundant after step \ding{194}-4.
We use proof by contradiction to verify this statement.
Let's assume that a token $V^n[i,j]$ is not selected by $M_p$ and still spatially redundant after step \ding{194}-4.
We can draw a \textbf{corollary} that its similarity to some spatially adjacent token $V^n[i+\delta_i,j+\delta_j]$ (with $(\delta_i, \delta_j)\in\{(-1,0),(1,0),(0,-1),(0,1)\}$) must exceed $\tau_s$ and $V^n[i+\delta_i,j+\delta_j]$ is also retained after step \ding{194}-4.
Since the token $(i,j)$ is not selected by $M_p$, we have
\begin{equation}
\label{eq:prove}
\begin{aligned}
    &\Big((i+\delta_i)+(j+(\delta_j)\Big)\%2 \\
  = & \Big((i+j)\%2+(\delta_i+\delta_j)\%2\Big)\%2 \\
  = & (0+1)\%2\\
  = & 1.
\end{aligned}
\end{equation}
According to equation (\ref{eq:mask}), the token $(i+\delta_i,j+\delta_j)$ must be selected by the mask.
Furthermore, according to equation (\ref{eq:spatial_redundancy}), $R_{i+\delta_i, j+\delta_j}\ge Sim(T_{i,j}, T_{i+\delta_i, j+\delta_j})\ge \tau_s$.
According to equation (\ref{eq:mns}), the token $(i+\delta_i,j+\delta_j)$  must be pruned in step \ding{194}-4, contradicting with the previous \textbf{corollary}.

\begin{table}[t]
\footnotesize
  \caption{The average performance over all the benchmarks.
The best and second-best scores are shown in bold and underlined, respectively.
\label{tab:exp-avg}}
    \setlength{\tabcolsep}{0.4em}
    \setlength{\abovecaptionskip}{6pt}
    \begin{center}
    
    \renewcommand{\arraystretch}{1.2}
    \begin{tabular}{ccc|cc}
    \toprule
    \multirow{2}{*}{\begin{tabular}[c]{@{}c@{}}\textbf{Method}\end{tabular}} & \multirow{2}{*}{\begin{tabular}[c]{@{}c@{}}\textbf{\#Frames}\end{tabular}} & \multirow{2}{*}{\begin{tabular}[c]{@{}c@{}}\textbf{Dropping}\\\textbf{Ratio}\end{tabular}} & \multicolumn{2}{c}{\textbf{Score}} \\ \cline{4-5}  
    & & & Absolute & Relative \\
    \midrule
      \specialrule{0em}{0.5pt}{0.5pt}
        
    TimeChat-Online-7B~\cite{timechatonline} & 1 fps & 0\% & 59.83 & 100\% \\ \hline
    \rowcolor{gray0}
    DTD~\cite{timechatonline} & 1 fps & 92.3\% & 56.17 & 93.88\% \\ \hline
    \rowcolor{gray0}
    VisionZip~\cite{yang2025visionzip} & 1 fps & 92.5\%  & 56.51 & 94.45\% \\ \hline
    \rowcolor{gray0}
    DivPrune~\cite{divprune} & 1 fps & 92.5\%  & \underline{57.01}& \underline{95.29\%} \\ \hline
    \rowcolor{gray0}
    \system & 1 fps & 92.5\% & \textbf{57.35} & \textbf{95.85\%} \\ \hline
    \rowcolor{gray2}
    DTD~\cite{timechatonline} & 1 fps & 93.3\% & 55.05 & 92.01\% \\ \hline
    \rowcolor{gray2}
    VisionZip~\cite{yang2025visionzip} & 1 fps & 93.5\% &  55.59 & 92.91\% \\ \hline
    \rowcolor{gray2}
    DivPrune~\cite{divprune} & 1 fps & 93.5\%  & \underline{56.38} & \underline{94.23\%} \\ \hline
    \rowcolor{gray2}
    \system & 1 fps & 93.5\% & \textbf{57.00} & \textbf{95.27\%} \\
    
    \bottomrule 
    \end{tabular}
    \end{center}
\end{table}

%% file: sec/4_experiment.tex
\section{Experimental Results}
\label{sec:exp}

\begin{table*}[t]
\footnotesize
  \caption{The experimental results on the real-time visual understanding part of the StreamingBench benchmark.
  The benchmark contains many tasks, including Object Perception (OP), Causal Reasoning (CR), Clips Summarization (CS), Attribute Perception (ATP), Event Understanding (EU), Text-Rich Understanding (TR), Prospective Reasoning (PR), Spatial Understanding (SU), Action Perception (ACP), and Counting (CT).
The best and the second-best scores are shown in bold and underlined, respectively.
  \label{tab:streamingbench}}
    \setlength{\tabcolsep}{0.4em}
    \setlength{\abovecaptionskip}{6pt}
    \begin{center}
    
    \renewcommand{\arraystretch}{1.2}
    \begin{tabular}{cc|cccccccccc|c}
    \toprule
    \textbf{Method} & \textbf{Dropping Ratio} & \textbf{OP} & \textbf{CR} & \textbf{CS} & \textbf{ATP} & \textbf{EU} & \textbf{TR} & \textbf{PR} & \textbf{SU} & \textbf{ACP} & \textbf{CT} & \textbf{All} \\ 
    \midrule
      \specialrule{0em}{0.5pt}{0.5pt}
        \midrule
    No Dropping &  0\% & 80.22  & 82.03 &  79.50 &  83.33 &  76.10 &  78.50 &  78.70  & 64.63  & 69.60  & 57.98 &  75.36 \\ \hline    
    \rowcolor{gray0}
    DTD~\cite{timechatonline} &  91.2\% & 73.98 & 78.91 & 75.08 & 73.40 &  67.30 & 68.22 &  75.00 & 59.35 & 67.33 &  27.13 & 67.28 \\ \hline
    \rowcolor{gray0}
    VisionZip~\cite{yang2025visionzip} &  91.5\%  & 74.25 & 82.03 & 76.97 & 76.60 & 72.96 & 73.21 & 73.15 & 60.16 & 68.75 & 32.98 & \underline{69.76} \\ \hline
    \rowcolor{gray0}
    DivPrune~\cite{divprune} &  91.5\%  & 77.78 & 78.91 & 75.71 &  77.24 & 70.44 & 72.27 & 75.00 & 60.16 & 65.62 &  27.13 & 68.96 \\ \hline
    \rowcolor{gray0}
    \system &  91.5\% & 78.05 & 79.69 & 76.97 & 75.64 & 72.96 & 72.59 & 80.56 & 63.01 &  67.05 & 31.38 & \textbf{70.24} \\ \hline
    \rowcolor{gray2}
    DTD~\cite{timechatonline} &  92.7\% & 71.27 & 81.25 & 73.19 & 72.44 & 67.30 & 66.67 &  69.44 & 59.76 & 63.64 & 19.68 & 65.16 \\\hline
    \rowcolor{gray2}
    VisionZip~\cite{yang2025visionzip} &  92.8\% &  73.98 & 79.69 & 76.97 & 74.68 & 69.81 & 71.65 & 73.15 & 58.54 & 66.76 & 30.32 & 68.32 \\ \hline
    
    \rowcolor{gray2}
    DivPrune~\cite{divprune} &  92.8\%  & 77.51 & 78.91 & 76.34 & 75.96 & 69.81 & 71.96 &  75.00 &  58.54 &  66.48 &  23.94 & \underline{68.48} \\ \hline
    \rowcolor{gray2}
    \system &  92.8\% & 75.88 & 80.47 &  76.03 & 76.60 & 69.81 & 70.40 & 80.56 & 60.57 & 66.76 & 30.85 & \textbf{69.16} \\
    
    \bottomrule 
    \end{tabular}
    \end{center}
\end{table*}

\begin{table*}[t]
\footnotesize
  \caption{The experimental results on the OVO-Bench benchmark.
  The benchmark contains many tasks, including Optical Character Recognition (OCR), Action Recognition (ACR), Attribute Recognition (ATR), Spatial Understanding (STU), Future Prediction (FPD), Object Recognition (OJR), Episodic Memory (EPM), Action Sequence Identification (ASI), Hallucination Detection (HLD), Repetition Event Count (REC),  Sequential Steps Recognition (SSR), and Clues Reveal Responding (CRR).
  \label{tab:ovobench}}
    \setlength{\tabcolsep}{0.4em}
    \setlength{\abovecaptionskip}{6pt}
    \begin{center}
    \renewcommand{\arraystretch}{1.2}
    \begin{tabular}{cc|cccccccccccc|c}
    \toprule
    \textbf{Method} &  \textbf{Dropping Ratio} & \textbf{OCR} & \textbf{ACR} & \textbf{ATR} & \textbf{ STU} & \textbf{FPD} & \textbf{OJR} & \textbf{EPM} & \textbf{ASI} & \textbf{HLD} & \textbf{REC} &  \textbf{SSR} & \textbf{CRR} & \textbf{All} \\ \midrule
      \specialrule{0em}{0.5pt}{0.5pt}
        \midrule

    No Dropping &  0\% & 75.2 &  46.8 &  70.7 &  47.8  & 69.3 &  61.4 & 55.9 &  59.5  & 9.7  & 31.6 &  38.5 &  40.0  & 46.7 \\ \hline
    \rowcolor{gray0}
    DTD~\cite{timechatonline} &  93.0\% & 65.77 & 43.12 & 55.17 & 41.57 & 64.36 & 52.72 & 49.83 & 56.76 & 9.68 & 27.45 & 33.63 & 40.00 & 42.08 \\ \hline
    \rowcolor{gray0}
    VisionZip~\cite{yang2025visionzip} &  93.0\% & 63.76 & 44.04 & 55.17 & 42.70 & 64.36 & 56.52 & 46.46 & 58.78 & 10.75 & 30.01 & 36.73 & 49.58 & \textbf{43.96}  \\ \hline
    \rowcolor{gray0}
    DivPrune~\cite{divprune} &  93.0\% & 63.76 & 45.87 & 61.21 & 46.63 & 69.31 & 55.43 & 47.14 & 60.14 & 6.99 & 30.75 & 36.03 & 42.92 & \underline{43.90}  \\ \hline
    \rowcolor{gray0}
    \system &  93.0\% & 69.13 & 44.04 & 62.07 & 43.26 & 66.34 & 53.80 & 51.85 &  56.76 & 6.99 & 30.11 & 34.73 & 41.67 & 43.49 \\ \hline
    \rowcolor{gray2}
    DTD~\cite{timechatonline} &  93.9\% & 56.38 & 40.37 & 55.17 & 42.13 & 61.39 & 52.17 & 48.15 & 57.43 & 9.68 & 24.89 & 33.25 & 40.42 & 40.85 \\\hline
    \rowcolor{gray2}
    VisionZip~\cite{yang2025visionzip} &  94.0\% & 60.40 & 42.20 & 56.90 & 39.33 & 66.34 &  54.35 &  44.78 & 57.43 &  9.14 & 29.75 &  36.66 & 49.58 & 43.01   \\ \hline
    \rowcolor{gray2}
    DivPrune~\cite{divprune} &  94.0\% & 63.09 & 44.04 & 65.52 & 44.38 & 69.31 & 54.35 & 48.48 & 58.78 &  6.45 &  30.66 & 35.70 & 44.58 & \textbf{43.89}   \\ \hline
    \rowcolor{gray2}
    \system &  94.0\% & 68.46 & 43.12 & 62.07 & 43.82 & 65.35 &  53.80 & 51.18 & 57.43 &  6.99 &  30.33 & 35.43 & 41.67 & \underline{43.48} \\
    \bottomrule 
    \end{tabular}
    \end{center}
\end{table*}

\begin{table}[t]
\footnotesize
  \caption{The experimental results on the VideoMME benchmark.
  The range of video duration is from 1 minute to 60 minutes, while the long part contains videos longer than 30 minutes.
  \label{tab:videomme}}
    \setlength{\tabcolsep}{0.4em}
    \setlength{\abovecaptionskip}{6pt}
    \begin{center}
    \renewcommand{\arraystretch}{1.2}
    \begin{tabular}{cc|cc}
    \toprule
    \multirow{2}{*}{\begin{tabular}[c]{@{}c@{}}\textbf{Method}\end{tabular}}
      & \multirow{2}{*}{\begin{tabular}[c]{@{}c@{}}\textbf{Dropping}\\\textbf{Ratio}\end{tabular}} & \multicolumn{2}{c}{\textbf{Video-MME}} \\ \cline{3-4} 
     &  &  overall & long \\
     
        \midrule
      \specialrule{0em}{0.5pt}{0.5pt}
        \midrule
        
      No Dropping  & 0\% & 62.4 & 48.4\\ \hline
      \rowcolor{gray0}
    DTD~\cite{timechatonline} &  92.2\% & \textbf{62.0} & \textbf{53.3} \\ \hline
    \rowcolor{gray0}
    VisionZip~\cite{yang2025visionzip} &  92.5\%  & 60.6 & 51.4  \\ \hline
    
    \rowcolor{gray0}
    DivPrune~\cite{divprune} &  92.5\%  & 61.3 & 51.9 \\ \hline
    \rowcolor{gray0}
    \system &  92.5\%  & \underline{61.6} & \underline{53.1} \\ \hline
      \rowcolor{gray2}
    DTD~\cite{timechatonline} &  93.0\% & \textbf{61.8} & \textbf{53.2} \\ \hline
    \rowcolor{gray2}
    VisionZip~\cite{yang2025visionzip} &  93.6\% & 60.0 & 51.7 \\ \hline
    
    \rowcolor{gray2}
    DivPrune~\cite{divprune} &  93.6\% & 60.3 & 52.1 \\ \hline
    \rowcolor{gray2}
    \system & 93.6\% & \underline{61.3} & \underline{53.0} \\ 
    \bottomrule 
    \end{tabular}
    \end{center}
\end{table}

\begin{table}[t]
\footnotesize
  \caption{The experimental results on LongVideoBench.
  \label{tab:LongVideoBench}}
    \setlength{\tabcolsep}{0.4em}
    \setlength{\abovecaptionskip}{6pt}
    \begin{center}
    \renewcommand{\arraystretch}{1.2}
    \begin{tabular}{ccc}
    \toprule
    \textbf{Method}  & \textbf{Dropping Ratio} & \textbf{Score} \\
        \midrule
      \specialrule{0em}{0.5pt}{0.5pt}
        \midrule
        
      No Dropping  & 0\% & 55.4\\ \hline
      
   \rowcolor{gray0}
    DTD~\cite{timechatonline} &  93.5\%  & 55.5 \\ \hline
    \rowcolor{gray0}
    VisionZip~\cite{yang2025visionzip} &  93.6\%  & 52.0 \\ \hline
    \rowcolor{gray0}
    DivPrune~\cite{divprune} &  93.6\%  & \textbf{55.7} \\ \hline
    \rowcolor{gray0}
    \system & 93.6\%  & \underline{56.2} \\ \hline
      \rowcolor{gray2}
    DTD~\cite{timechatonline} &  94.6\% &  \underline{54.8} \\ \hline
    \rowcolor{gray2}
    VisionZip~\cite{yang2025visionzip} &  94.7\% & 51.5  \\ \hline
    \rowcolor{gray2}
    DivPrune~\cite{divprune} &  94.7\% & 54.3  \\ \hline
    \rowcolor{gray2}
    \system & 94.7\% & \textbf{56.2} \\ 
    \bottomrule 
    \end{tabular}
    \end{center}
\end{table}

\subsection{Experimental Settings}
\label{sec:exp-settings}

\para{Settings of \system.}
We evaluate \system with the open-source TimeChat-Online-7B model~\cite{timechatonline}.
This model is built by finetuning the Qwen2.5-VL-7B model to adapt to pruned input of video tokens.
We set the temporal dropping threshold $\tau_t$ as $0.2$ and the spatial dropping threshold $\tau_s$ as $0.2$ or $0.5$.
We process the video at 1fps for evaluation.

\para{Settings of baselines.}
We use a baseline of no dropping as well as many baseline token pruning methods.
During evaluation, we use the same open-source online video LLM model TimeChat-Online-7B~\cite{timechatonline}).
The evaluated token pruning methods include DTD~\cite{timechatonline}, VisionZip~\cite{yang2025visionzip}, and DivPrune~\cite{divprune}.
DTD seeks to pinpoint temporally redundant tokens, while VisionZip and DivPrune pinpoint spatially redundant tokens through an attention-based strategy and a similarity-based strategy, respectively.
For evaluation of the DTD baseline, we cannot directly control its dropping ratios and thus adjust its temporal dropping threshold for multiple times so that the resulting dropping ratio is close to that of \system.
For evaluation of the VisionZip and DivPrune baseline, we use an equal pruning ratio for every frame in any single video and set each video's pruning ratio to the same as that of \system.
We also follow the setting in the paper of VisionZip by controlling the number of dominant visual tokens to 5.4 times more than that of contextual visual tokens.

\para{Metrics.}
We aim to improve the model's inference latency while preserving the response quality with token pruning.
However, we mainly report the accuracy under different pruning ratios.
We mainly use pruning ratios rather than latencies for two reasons.
\First, \system and the baseline pruning methods all apply token pruning for every incoming frame independently and incur a negligible pre-processing latency (\ie less than 1 millisecond).
Therefore, different pruning methods incur almost the same TTFT latency when the pruning ratio is the same.
\Second, even if the pruning ratio is fixed, the inference latency is not constant and can be influenced by factors such as GPU capacity.

\subsection{Results on Video Benchmarks}
\label{sec:exp-streaming}
We evaluate \system on both the streaming video benchmarks (including StreamingBench~\cite{StreamingBench} and OVO-Bench~\cite{ovo-bench}) and offline video benchmarks (including Video-MME~\cite{videomme} and LongVideoBench~\cite{wu2024longvideobench}).
In streaming video benchmarks, each question is paired with a timestamp attribute, and we only use the video part before the timestamp during evaluation.
As for offline video benchmarks, we directly use the entire video for evaluation.
The four benchmarks contain more than 9000 samples in total and provide comprehensive evaluation results.

The average results are shown in Table~\ref {tab:exp-avg}.
The separate results for each benchmark are shown in Table.~\ref{tab:streamingbench},~\ref{tab:ovobench},~\ref{tab:videomme}, and~\ref{tab:LongVideoBench}, respectively.
We have several key observations from the results.
\First, \system outperforms all the baselines on average.
Its performance on each benchmark is also very high, demonstrating a high generalizability.
\Second, the accuracy of \system is rather close to the strategy of no dropping even at a high pruning ratio.
Even if as many as 93.5\% of tokens are pruned, Streaming-Assistant still maintains a high relative accuracy score of 95.27\%.
Overall, the experimental results demonstrate the huge efficiency advantages of \system.

\subsection{Ablation Studies}
\label{sec:exp-ablation}

\para{Variants.}
We conduct comprehensive ablation studies to demonstrate the effectiveness of \system's key designs.
Specifically, we evaluate several variants of the proposed spatial pruning method.
The first variant \system-IA (\textbf{I}nterleaving \& \textbf{A}djacent) uses the strategy of interleaving pruning and redundancy update and still quantifies the spatial redundancy through the MSSAVT metric.
This variant corresponds to the sequential pruning strategy in Fig.~\ref{fig:workflow}.
The second variant \system-IG (\textbf{I}nterleaving \& \textbf{G}lobal) uses the strategy of interleaving pruning and redundancy update and quantifies the spatial redundancy as the maximum to any other token within the entire frame.
The third variant \system-NA (\textbf{N}on-interleaving \& \textbf{A}djacent) quantifies the spatial redundancy through the MSSAVT metric but uses the naive pruning strategy of computing the spatial dropping mask $M_s^n$ through $R_s>\tau_s$.
This variant corresponds to the naive parallel strategy in Fig.~\ref{fig:workflow}.

We fix $\tau_t$ to $0.6$ and vary $\tau_s$ to obtain each variant's performance (including both the accuracy and the latency) under different pruning ratios.
As each benchmark contains many samples and each sample has its own pruning ratio, we report the average pruning ratios over all the samples.
We use StreamingBench  for ablation studies.

\begin{figure}[tp]
    \centering
    \includegraphics[width=\linewidth]{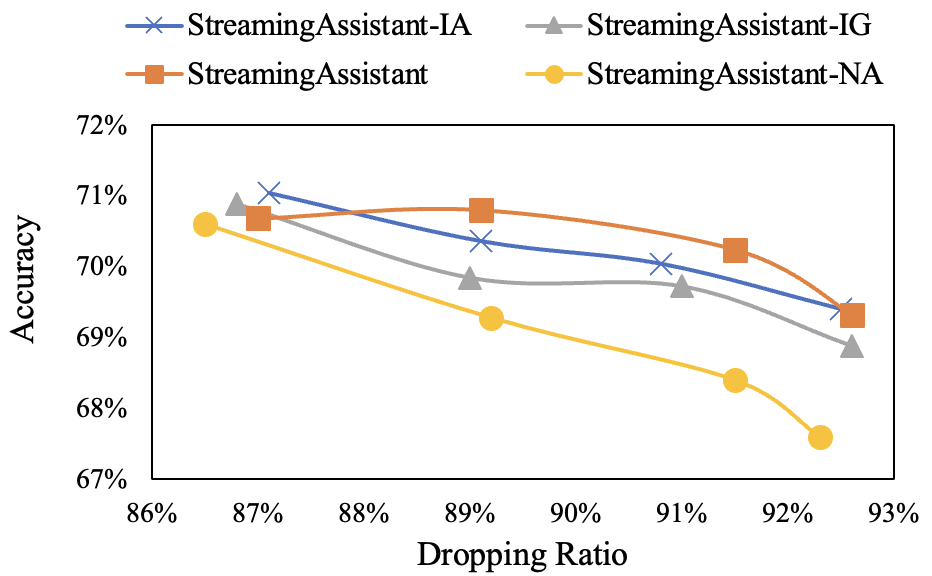}
   \caption{The accuracy results of the ablation studies.}
    \label{fig:ablation-acc}
\end{figure}

\para{Accuracy results.}
As shown in Fig.~\ref{fig:ablation-acc}, \system-NA has the worst accuracy under the same level of pruning.
This demonstrates that the naive pruning strategy ignores the entanglement between pruning and redundancy quantification and thus can lead to unreliable pruning results.
In contrast, both \system and \system-IA have significantly higher accuracies as they consider the entanglement issue.

Furthermore, the performance of \system-IG is slightly worse than \system and \system-IA.
The reason is that \system-IG computes the redundancy in a larger spatial range and may prune a token if it is similar to a remote token.
However, a pair of spatially distant yet similarity-valued tokens may be better retained because they share very different positional information.
This demonstrates the advantage of the  MSSAVT metric.

\para{Latency results.}
Although \system has a comparable accuracy to \system-IA, its pruning efficiency is significantly higher.
To quantify the advantage, we measure the pruning latency of \system and \system-IA on an idle NVIDIA A100 GPU with 80GB memory.
As \system and \system-IA prune video tokens from every incoming frame independently, the reported pruning latency is the latency for processing a single frame.
The reported pruning latency is defined as the duration from the encoder generating all the video tokens of a single frame to both the temporal and spatial pruning modules finishing processing.

\begin{figure}[tp]
    \centering
    \includegraphics[width=0.8\linewidth]{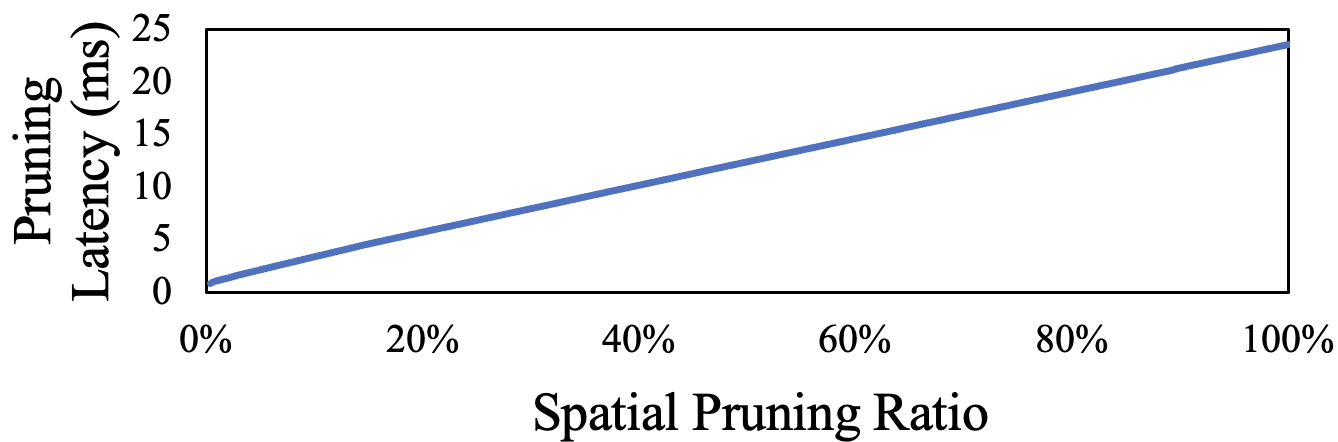}
   \caption{The pruning latency of \system-IA under different spatial pruning ratios.}
    \label{fig:ablation-latency}
\end{figure}

The pruning latency of \system is always less than 1 ms and agnostic to the pruning ratio.
In contrast, as shown in Fig.~\ref{fig:ablation-latency}, the pruning latency of \system-IA variant increases linearly with the spatial dropping ratio and reaches a maximum latency of $23.6$ ms for the StreamingBench benchmark.
Such a high pruning latency prohibits its usage in many interactive and latency-sensitive applications~\cite{wu2025chataisurprisingturn}.
Furthermore, each video frame in the benchmark is downscaled to have fewer than $448\times448$ pixels and then transformed to only $252$ tokens.
In applications requiring a high video resolution (such as the 4K resolution of 3840 x 2160 pixels), the maximum pruning latency can be as large as about 1 second, which is comparable to the LLM part's TTFT latency under an input sequence length of more than ten thousand tokens.

%% file: sec/5_future_work.tex
\section{Discussion}
\label{sec:future-work}

\para{Advanced buffer control.}
To enhance \system's practicality for real-world scenarios where late user queries can cause the buffer to overflow, we plan to design an automatic buffer update strategy. This strategy will store each video token with its corresponding redundancy value and evict the most redundant tokens once the buffer is full, thereby preventing out-of-memory errors and managing latency. Furthermore, this design will enable flexible control over response time: given a latency requirement, we can select the least redundant tokens from the buffer to meet the target. This approach will ensure \system remains robust and efficient under long-duration videos.

%% file: sec/6_conclusion.tex
\section{Conclusion}
\label{sec:conclusion}
In this paper, we introduced \system, an efficient token pruning method for online video understanding with MLLMs.
To address the limitations of existing methods, we propose the MSSAVT metric, which accurately quantifies spatial redundancy while preserving crucial positional information.
We further identify and resolve the critical entanglement between redundancy quantification and pruning through a theoretically sound masked strategy, enabling fast and parallel processing.
Extensive experiments confirm that \system achieves superior accuracy over existing baselines with negligible pre-processing latency, making it a highly effective solution for online video understanding.